# Scene recognition based on DNN and game theory with its applications in human-robot interaction

R.Q. Wang[1*], W.Z. Wang[1], D.Z. Zhao[1], G.H. Chen[1] and D.S Luo[2]

1. College of Mechanical and Electrical Engineering, Beijing University of Chemical Technology

2. Department of Machine Intelligent, Peking University

**Abstract.** Scene recognition model based on the DNN and game theory with its applications in human-robot interaction is proposed in this paper. The use of deep learning methods in the field of scene recognition is still in its infancy, but has become an important trend in the future. As the innovative idea of the paper, we propose the following novelties. (1) In this paper, the image registration problem is transformed into a problem of minimum energy in Markov Random Field to finalize the image pre-processing task. Game theory is used to find the optimal. (2) We select neighboring homogeneous sample features and the neighboring heterogeneous sample features for the extracted sample features to build a triple and modify the traditional neural network to propose the novel DNN for scene understanding. (3) The robot control is well combined to guide the robot vision for multiple tasks. The experiment is then conducted to validate the overall performance.

**Keywords:** Scene recognition; DNN; Game theory; human-robot interaction; Machine learning

## 1    Introduction

Motion vision analysis uses computer technology to detect, classify, and track people from image sequences containing people, and to understand and describe their behavior. The characteristic extraction and the movement attribute are in to first floors and in the intermediate deck processing and so on goal examination, classification and track foundation, withdraws the picture target characteristic from the goal movement information and uses for to attribute the target travel condition; the behavior recognition is inputs the movement characteristic and the reference sequence which in the sequence withdraws carries on the match, which behavior model judges the current movement to be in; the high-level behavior and the scene understanding are unify the scene information and the related domain knowledge which the behavior occurs, the recognition complex behavior which realizes to the event and the scene understanding [1, 2, 3]. Traditional scene recognition methods generally use low-level features or the high-level features. The advantages of these methods are simple and easy to implement, and they also have good logic and are consistent with human intuitive perception [4, 5]. But when the data to be processed reaches a certain scale and the scene classification reaches a certain amount, the traditional low-level features and high-level features cannot represent so much scene information, and deep learning-based methods are very suitable for dealing with such problems [6, 7]. The use of deep learning methods in the field of image scene recognition is still in its infancy, but has become an important trend in the future [8, 9, 10].

At the same time, the image capture methodologies are also essential for the overall analysis. Currently, most 3D datasets used to test the 3D target recognition algorithms are obtained under specific scenarios, from a single perspective, few target categories, and a single instance. These 3D data test sets are only used in specific scenarios and lack a wide range of the practical applications. The specific data set obtained by the Kinect sensor can contain a large number of common objects in daily life. It is the data obtained by shooting from multiple perspectives in indoor or office environments, and can be used for the large-scale 3D object recognition practical application, which is more practical than previous data sets. Kinect makes full use of the PrimeSense's PrimeSense device and sensor chip PS1080 from Israel to realize the depth camera function at a low cost [11, 12, 13]. The information provided by the depth camera, RGB camera, and multi-array microphone enables the Xbox to sense and recognize players and their movements, thereby manipulating the characters in the game and improving the playability of the game. Not only Kinect may play the game, moreover because it provided the performance-to-price ratio very high depth camera function to obtain the world each place research and development personnel's attention, therefore Kinect widely is applied in each kind of research area specially, especially is active in the computer vision and the robot study domain. Deep learning based, Kinect assisted vision analysis and scene understanding model is the trend.

For the scene understanding, there are three major challenges. (1) Pedestrian attributes have complex local characteristics, which mean that some attributes can only be identified in certain identified or uncertain local body regions. (2) Pedestrian attribute classification is the multi-label classification problem rather than a multi-category classification problem, because the pedestrian attributes are not mutually exclusive [14, 15, 16]. (3) Due to the variety of clothing appearance, the difference of lighting conditions and camera angle of view, serious intra class differences are caused.

Deep learning gives this question a guiding star. The deep learning-based scene recognition system is mainly divided into front-end construction of local descriptors and the back-end optimization and loop detection of constructed map. Improvements to the front-end work include the use of deep learning-based local descriptor camera poses to directly calculate, while in summary, the target acceleration feature points are matched [17, 18]. The front end mainly provides the extraction of feature points, the calculation of camera pose and the construction of local maps. In the Figure 1, we give the simple demonstration of the deep neural networks (DNN).

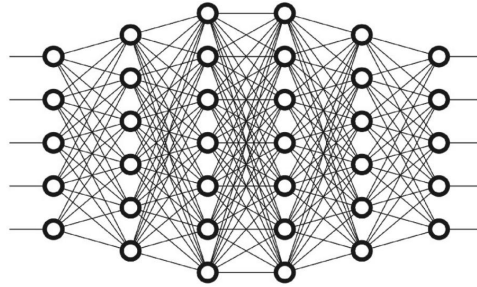

**Fig.1.** The Simple Demonstration of the Deep Neural Networks

Deep neural network, as a classic model of deep learning can automatically learn and extract features from data, and its generalization ability is significantly better than traditional methods. In the Table 1, we define the procedures of the DNN.

**Table 1.** The Principles Procedures of the DNN

| Procedures | Description |
|---|---|
| Convolution operation | The neural network essence is the affine transformation operation: The input vector, with a matrix multiplication, its result takes the network the output that suits in processes any data type the question, no matter is the image, the sound, perhaps the random characteristic set, their characteristic all may stretch is a vector. The convolution layer is a linear operation unit that can effectively preserve the structural properties of image data during the forward propagation of the target image data [19, 20, 21]. |
| Pooling operation | The pooling operation usually average the down-sampling or maximum down-sampling of the input feature map. In the convolution neural network, the pooling operation may reduce the characteristic chart size and reduce the computation order of complexity. On the other hand, it may promote the network to study that has the translation invariable scene characteristic. |
| Activation operation | The activation function is an important part of the convolutional neural network. It is usually used to increase the nonlinearity of the network and improve the expression ability of the model. |
| Connection operation | In the deep neural network, the full connection is generally placed at the end of the convolution neural network, which acts as the classifier, mapping the high-level convolution features acquired to the sample marker space. |

After the scene image passes through multi-scale layers, scene image blocks of the multiple scales are obtained. Scene blocks of the different scales are input to different sub-convolutional networks for high-level semantic feature extraction. On the other hand, deep learning adopts the feature pre-fusion strategy, and uses the in-scale feature fusion module to fuse scene block features at the same scale in each sub-convolution network. For the efficient processing of the images, the pre-processing operation is essential. In this manuscript, we consider using the game theory to finish the task. Because there is no universal algorithm for image enhancement technology, image enhancement technology has produced a variety of algorithms according to various purposes. The most commonly used methods are the "spatial domain" method and the "frequency domain" method. Along with mathematics various branches in the theory and application gradually thorough, caused mathematics morphology, the fuzzy mathematics, the genetic algorithm, the wavelet theory and so on to make the very big progress in the image intensification technology application, there has had many new algorithms. For the models, listed key factors should be considered. (1) The enhanced image should have good visual effects. Avoid excessive or weak local enhancement of the enhanced image. The enhanced image should meet the visual characteristics of the human eye. (2) The image enhancement algorithm should have better real-time performance. With the rapid development of embedded products in recent years, the real-time requirement of image enhancement algorithm is more and more. (3) During the enhancement image step, should avoid enlarging the noise. If cannot the noise be processed as elimination, the noise be able to enlarge effectively in the image intensification process, thus creates the shade to the picture quality [22].

Game theory is used to deal with parallel decision-making problems in a conflict environment. It decouples complex systems into modules, and each module makes distributed decisions based on information interaction. It has a strict mathematical theoretical foundation. Taking the case of the two characteristics as an example, this article considers the two characteristics as two players in the round, and completes the registration through the non-zero and non-cooperative games between people in the round. In gambling, two players both attempt to cause own income maximization or the price minimum, therefore two kinds of characteristics are the decouplings. The gambling matching frame balanced solution obtains by the rational decision-making, and this ratio optimizes a simple target function to have generality. Therefore, with the use of the game theory, the pre-processing will be efficient [23, 24].

In the rest of the paper, we discuss the proposed methodology. In the section 2, the proposed model is analyzed. In the section 3, the experiment is conducted. In the section 4, the conclusion is presented. To begin with, in the Figure 2, we present he data set used for the paper.

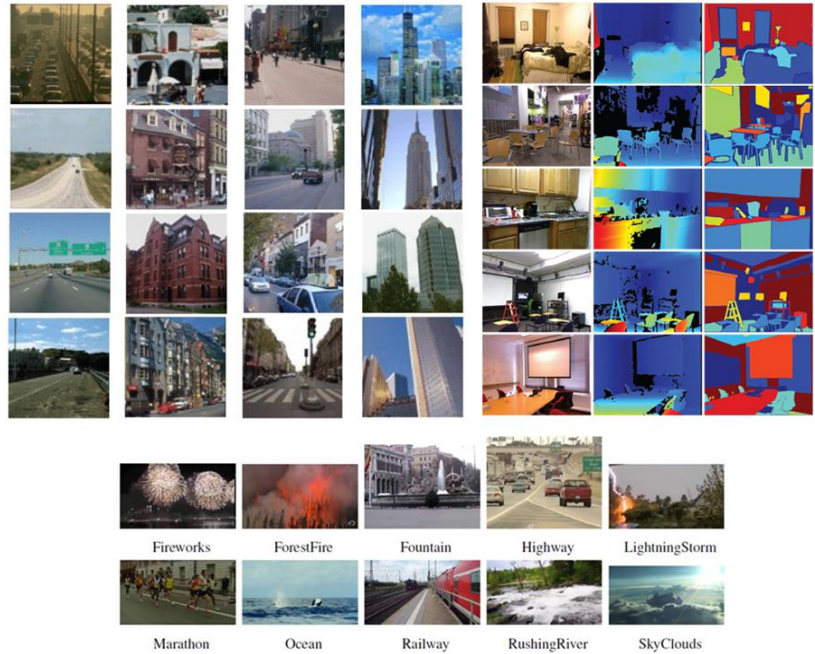
**Fig.2.** The Dataset Used for the Manuscript

## 2  The Proposed Methodology

### 2.1  The Image Pre-processing

**2.1.1 Review of the Image Pre-processing Models**

   Generally speaking, the images people see are actually not real images, but just the simulated images that are felt by the brain and different colors on the images will also form the different shapes. When a person watches a specific thing, if it is a moving object, the image seen will have different colors and shapes, which can be used as the function of the image space coordinates to calculate. Collecting images is generally completed by three steps: imaging, sampling, and quantization. Spatial coordinates of the image are obtained by sampling. Because time photography image, the object reflected light is not 100% parallel light, simultaneously light brightness also has the non-uniformity that adds the outside light disturbance as well as the object surface factor influence, the image can become fuzzy. Also some one kind of situation also possibly causes the image the degeneration, that is the transformation when arrives likely by the mapping the digital image, this process is not artificial holds controls, possibly appears each kind of unwanted signal that has the possibility to have the deviation. For example, if the image shape is destroyed, or image itself is looks not clearly. Therefore, in order to guarantee the image also may examine in each kind of the meaconing situation that needs to carry on the essential adjustment to the image, simultaneously avoids the image useful signal also receiving the disturbance.

In the table 2, we review the current models for references [25, 26].

**Table 2.** The Image Pre-processing Models

| Model | Description |
|---|---|
| Frequency domain image enhancement | Fourier transform is applied to the image to obtain its spectrum, the zero frequency components is equal to the average grayscale of the image, the smooth image signal contributes the low-frequency component in the frequency domain, the details and boundaries in the image contribute to the partition of the high-frequency domain, and the noise spectrum has rich high-frequency components. |
| Histogram equalization algorithm | Histogram algorithm uses the cumulative distribution function to map the specified input gray level to the output gray level, so that the output gray level has an approximately uniformly distributed probability density function, so as to improve image contrast and enlarge purpose of image dynamic range |
| GAN based model | GAN, while improving the visual quality of the generated image model samples, has increased attention to the overfitting of the memory process of the training samples. GAN model can contain the pattern avalanche phenomenon effectively and the production image type is richer, not easy to appear the identical image. In the training process, through overlapping targets and so on entropy and rate of accuracy instructed training process, also GAN training better, the target value is smaller, the generator produces the picture quality is higher. |
| Wavelet transform based image enhancement | After the original image is decomposed into different frequency images by wavelet transform, the images with different frequency can be enhanced by various methods. After wavelet decomposition of the image, the low-frequency component of the image is enhanced by histogram equalization. The high-frequency component of the image is enhanced by maximizing the contrast entropy. Finally, the enhanced image is obtained through wavelet reconstruction. |

### 2.1.2 Game Theory for the Pre-processing

Game theory is an important branch of mathematics. It studies what kind of rational strategies should be adopted by the players in the environment of the multi-person confrontation or cooperation in order to obtain favorable personal interests or the collective interests of the alliance. In this paper, the discrete displacement field is used to represent the deformation. The registration problem is transformed into a problem of minimum energy in random field. The image registration is regarded as a multi-person non-cooperative game process between pixels. By finding Nash equilibrium point of the game to get minimum energy point and complete image preprocessing. For obtaining the optimal, we should use the EM model as follows.

| **Algorithm Flow**. The (EM) Model |
|---|
| 1. **Input**: The dataset (x), the total number of clusters (M), the accepted error for convergence (e) and the maximum number of iterations. |
| 2. Compute the expectation of the complete data log-likelihood. $$Q(\theta,\theta^T) = E\left[\log p(x^g,x^m\mid\theta)x^g,\theta^T\right]$$ |
| 3. Select a new parameter estimate that maximizes the Q-function. $$\theta^{T+1} = \arg\max_\theta Q(\theta,\theta^T)$$ |
| 4. Increment t=t+1; repeat steps 2 and 3 until the convergence condition. |
| 5. **Output**: A series of parameters represents the achievement of the convergence criterion. |

There are many kinds of energy optimization methods for the model. Among them, global optimization relaxation algorithm requires a lot of time to obtain the optimal solution. Although the local optimization relaxation algorithm has a fast processing speed, it can only obtain the local optimal solution and the segmentation effect which will be poor, while not suitable for actual image processing projects. For background segmentation, we can define the question as formula 1.

$$P(X\mid Y) = \frac{P(Y\mid X)P(X)}{P(Y)} \quad (1)$$

Where the $P(X\mid Y)$ is the target function and the $P(Y\mid X)P(X)$ is prior information. After each player chooses a strategy, a game situation is formed, and equilibrium is a special situation. When each player cannot unilaterally reduce the value of the cost function, the game ends, and the situation at this time is called the Nash equilibrium. In the formula 2, we define this scenario.

$$F1: P1\times P2 \to F2: P1\times P2 \to R_{optimal} \quad (2)$$

As the same time, the prior information is defined as formula 3.

$$P_{prior} = \frac{\exp U(X)}{z} \quad (3)$$

Where the $z$ is normalization constant, defined prior distribution is a probabilistic description of general knowledge of image structure, which indicates the possibility of deformation. Generally, the possibility of large deformation is small, while the possibility of small deformation is large. This paper uses MRF to describe the prior distribution. The model is defined as formula 4.

$$Obj = \sum_{i=1}^{n}\hat{u}_i^2 = \sum_{i=1}^{n}(y_i-\hat{y}_i)^2 = \sum_{i=1}^{n}[y_i-(\hat{a}+\hat{b}x_i)]^2 = \sum_{i=1}^{n}(y_i-\hat{a}-\hat{b}x_i)^2 \quad (4)$$

We consider the pixels as the people in the game, the pixel markings as the strategy of the people in the game, and the cost function as the energy function. Then the energy optimization problem is the game between pixels. Each pixel is based on Self-return function and neighbor's strategy, choose the strategy that maximizes self-return and minimize the total energy, that is, the Nash equilibrium point of the game is reached. In the formula 5, we define the data sets for the analysis

$$H(image) = [h(x_1), h(x_2), h(x_3), \cdots, h(x_n)] \tag{5}$$

The judgment condition is simple and the convergence speed is fast, but it is easy to fall into the local optimum. We use the following steps to find the optimal.

---

**Algorithm Flow**. The Improved Finding Optimal Algorithm

---

1. **Input**: The row dataset $O$, the clustering threshold $\theta$.
2. Calculate similarity between every pair of features and merge the two most similar features into one.
3. **Repeat:**
   i. Calculate similarity between every two clusters.
   ii. Merge the two most similar into one cluster.
4. **Until:**
   The maximum similarity is less than $\theta$.
5. Calculate F-Completeness of every feature and set S to be $\Phi$.
6. Add the features into S.
7. **Output**: Selected feature subset S.

---

It is a typical random relaxation law. This method can theoretically obtain the global result and will not fall into the local optimal solution. Computation is slow due to random perturbations. Therefore, we introduce the core object function as the formula 6.

$$\max \sum_{i=1}^{n} \left( \frac{\sum_{j=1}^{m} w_j \sum_{q=1}^{q_0} \left( H_{ij}^{\sigma(q)} - \left( H_j^{\sigma(q)} \right)^- \right)}{\sum_{j=1}^{m} w_j \sum_{q=1}^{q_0} \left( \left( H_j^{\sigma(q)} \right)^+ - \left( H_j^{\sigma(q)} \right)^- \right)} \right)$$

$$s.t. \quad w = (w_1, w_2, \ldots, w_m)^T \in \Delta \tag{6}$$

$$w_j \geq 0, \quad j = 1, 2, \ldots, m, \quad \sum_{j=1}^{m} w_j = 1.$$

Where the $\sum_{j=1}^{m} w_j \sum_{q=1}^{q_0} \left( H_{ij}^{\sigma(q)} - \left( H_j^{\sigma(q)} \right)^- \right)$ is the game initial information and the $\sum_{j=1}^{m} w_j \sum_{q=1}^{q_0} \left( \left( H_j^{\sigma(q)} \right)^+ - \left( H_j^{\sigma(q)} \right)^- \right)$ is the later function. If for any initial point, the Nash equilibrium exists, it is said that the Nash equilibrium is stable. If the equilibrium exists only when the initial point is within a certain neighborhood of the equilibrium, the Nash equilibrium is said to be locally stable. In our model, we will use the testing model to find the optimal solution.

## 2.2 The Proposed Scene Understanding Model

### 2.2.1 Review of the Scene Understanding Models

In human motion visual analysis, due to the influence of the perspective, the same behavior may have different projection trajectories, and different behaviors may have the same projection trajectory. In addition, the realistic environmental factors such as lighting changes and occlusions have also brought great difficulties in understanding behavior. Realizing the 3D target identification method based on the 2D image recognition algorithm including based on the depth chart HOG characteristic method, the machine learning method and the depth nuclear descriptor method and so on. These methods all are in the original 2D pattern recognition algorithm foundation, withdraws the characteristic based on the depth chart to carry on the recognition. Scene recognition is classified according to application scenarios, and can be divided into outdoor scene recognition and indoor scene recognition. According to the type of features used, it can be divided into four types of methods, namely, bottom feature methods, middle-level semantic methods, high-level feature methods, and learning feature method. Simple features are suitable for outdoor scene recognition with low complexity, but in some scenes with more foreground targets, it is difficult to achieve a good effect. It has gathered the prior model knowledge based on the model method and the current input, can suit the complex behavior the behavior understanding, but usually requests first to choose the model, initialization is difficult, the computation load is big, moreover as a result of the human movement high degree of freedom, easy to have the local minimum, very difficult to find the overall situation to be most superior also the robust model parameter, in addition as a result of at the following track turban existence accumulative error, it cannot analyze or track the long sequence movement. Feature extraction, as the most critical link in the 3D target recognition, largely determines the performance of the recognition system and features mainly include two types of global features and local features. The global feature has a small amount of calculation and is easy to implement. It is widely used in the 3D model retrieval. But it is not strong in discriminating details, and requires pre-segmentation of the target and complete 3D data. In addition, due to the simplicity of features, but the number of scene categories increases, the underlying features lack sufficient scene information for the scene classification and recognition. Inspired by this, in the table 3, we demonstrate the current scene understanding models.

Table 3. The Scene Understanding Models

| Model | Description |
| --- | --- |
| Hidden Markov and improved model | HMMs use Markov processes to establish the correlation between adjacent instances, and assume that the observation sequence is determined by a hidden process consisting of a fixed number of hidden states, which is a random state. HMMs is used as hybrid models of these intermediate power systems to express complex motions. The recognition process is completed by maximizing the posterior probability of HMMs. |
| Curvature-based model | It examines the characteristic point based on the curvature method using the sampling point and the neighborhood curvature information. This kind of method the characteristic point which obtains compared to the stochastic sampling law has better may duplicate the extractability, the validity and the |

| | invariance under transformation, but the curvature computation request contains a denser spot from the image, also in curvature differentiate easily noise influence |
|---|---|
| Neural network based model | The large number of data sets is highly operable, and the focus of time-delay neural networks can be on the expression of time-division information. In this way, the neural network model derived from this is suitable for processing sequence data. The NN based method has the feature of collecting massive data to solve the complex issue and the training is efficient, robustness and accuracy can be guaranteed. |

### 2.2.2 DNN based Model

In machine learning, the combination method is to learn the multiple classifiers and combine them to form an efficient classifier. The combination method is also suitable for high-semantic image recognition tasks. Cross-validation uses different training data and validation data to train the network, which is a simple combination method. DNN was first developed by scientists in the 1960s by observing the visual cortex cells of cats and the concept of receptive fields, that is, each visual neuron will only process visual images of a small area, in the Figure 3, the initial model is shown.

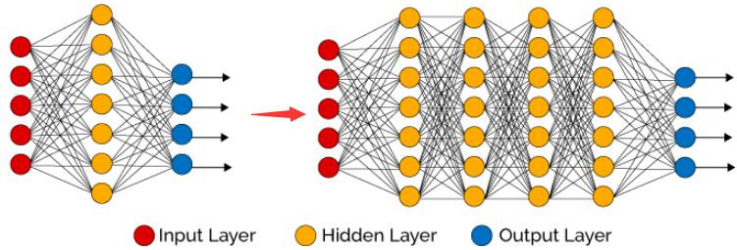

**Fig.3.** The Initial Model of the Deep Neural Networks

In our proposed model, the feature extraction is initial process. The characteristic extraction is in the computer vision research important link, in order to adapt the different duty and it chooses the appropriate characteristic to be often twice the result with half effort. The characteristic extraction is actually one kind of data compression process, the extraction characteristic already must guarantee the primary information greatest degree retains, which also must consider counting yield factors and so on characteristic dimension and extraction time. In the formula 7, we demonstrate the step of the feature projection.

$$\begin{bmatrix} z_1^0 & z_2^0 & \cdots & z_M^0 \\ z_1^1 & z_2^1 & \cdots & z_M^1 \\ \vdots & \vdots & & \vdots \\ z_1^{M-1} & z_2^{M-1} & \cdots & z_M^{M-1} \end{bmatrix} \cdot \begin{bmatrix} h_1 \\ h_2 \\ \vdots \\ h_M \end{bmatrix} = \begin{bmatrix} x[1] \\ x[2] \\ \vdots \\ x[M] \end{bmatrix} \qquad (7)$$

In the formula 7, the $z_M^N$ denotes the original data sets, and the $h_M$ is the operational function, the $x[M]$ presents the processed data matrix. Feature coding is result of counting a large number of local feature distributions. Studies have shown that there is a strong correlation between statistical distribution characteristics and categories of a large number of local features, and feature coding is a transformation bridge between local and global features of an image. With the data scale increases, it has brought some challenges for the computer vision duty, in the image match duty, has the possibility to match extremely looked resembles the similar actually actual irrelevant picture, and in picture recognition domain, because a kind of similar existence, it brings the disturbance to the recognition.

$$\min_d \nabla f(x^k)^T d + \frac{1}{2} d^T H_k d, \quad Subject-to$$
$$c(x^k) + \nabla c(x^k)^T d \leq 0,$$
$$h(x^k) + \nabla h(x^k)^T d = 0$$
(8)

To provide the precious description of the model, the formula 8 gives the optimal model and the restriction function, where the $\nabla f(x^k)^T d + \frac{1}{2} d^T H_k d$ is the objective function and the $c(x^k) + \nabla c(x^k)^T d \leq 0,$ $h(x^k) + \nabla h(x^k)^T d = 0$ is boundary parameters. DNN combined with some large-scale data sets has achieved satisfactory results in the object recognition applications, but scene recognition has long relied on traditional features or fine-tuning based on object recognition models, which is not enough. Therefore, we should design the efficient model for analysis. In order to meet the real-time requirements of the closed-loop detection algorithms, we design the fast and streamlined deep neural network model for closed-loop detection shown in the Figure 4.

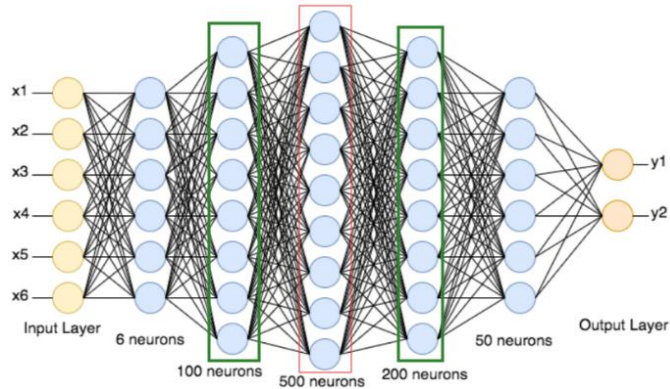

**Fig.4.** The Designed DNN Structure

For the proposed model, the loss function is also essential. We select neighboring homogeneous sample features and neighboring heterogeneous sample features for the extracted sample features to build a triple. For N training samples, a large number of triples can be randomly generated. Tuple-constrained distance metric learning models require that the distances of features of the same sample and the distances of features of the different sample be separated by a large interval as formula 9.

$$\sum_{i=1}^{n}\sum_{j=1}^{n}\frac{\partial}{\partial x_i}\{a_{i,j}(x)\frac{\partial}{\partial x_i}u(x)\}=0 \qquad (9)$$

Where the $a_{i,j}(x)\frac{\partial}{\partial x_i}u(x)$ is inner parameter and the $\frac{\partial}{\partial x_i}$ is the further operation.

Existing research indicates that in the recognition task, large-scale training data, deep recognition models, and the application of avoiding the overfitting techniques are the directions to improve the recognition accuracy. The large data sets that have appeared in recent years have provided conditions for computer to learn further more optimized models. Considering this, we should pay special attention to 2 major aspects. (1) The pooling layer can reduce the size of the matrix, thereby reducing the parameters in the last fully connected layer. The use of pooling layers can speed up calculations and prevent overfitting. (2) The partial response normalization level, this level founded the competitive system to the partial neuron, causes in which to respond the great value to become bigger, and suppresses other feedback small neuron, enhancement model exudes ability.

Considering mentioned issues, we should revise objective function's optimization parameter as formula 10.

$$Minimize\ F(\bar{x}) = \sum_{i=1}^{n} a_i f_i(\bar{x}), \quad a_i > 0, \ i = 1, 2, \ldots, n \qquad (10)$$

Data augmentation is most effective way to avoid overfitting. Data transformation is one of the effective methods for data augmentation. It is to artificially make some changes to the original image while ensuring that the labels are not changed.

The $(\bar{x})$ is the target vector, and the $\sum_{i=1}^{n} a_i f_i(\bar{x})$ is the further option. In view of the different scene recognition technology existence question, we worked are for the purpose of improving formerly the depth mix recognition frame, integrated the depth from encoder automatic extraction partial image block characteristic, the introduction space information improvement scene picture expression method with perfect depth mix scene recognition frame. Therefore, we add new parameter as formula 11.

$$\sum_{i=1}^{n}\sum_{j=1}^{n} a_{i,j}(x)\xi_i\xi_j \geq \varepsilon \sum_{i=1}^{n} \xi_i^2 \qquad (11)$$

In the formula 11, the $\varepsilon$ is the adjusting parameter and the $\xi_i^2$ is the summary of the target objects. The $a_{i,j}(x)\xi_i\xi_j$ is the summary of the objects in the projected space.

In order to solve the problem of insufficient data volume, in addition to collecting more data, it is also a common method to augment existing data based on existing data to extract fixed-size sub-regions from the four corners and middle positions of the original image in image depth recognition as a new training picture. In the Figure 5, we demonstrate the complete framework for the proposed model. Firstly, the images are captured through the sensors as the source data, the probability estimation is analyzed in the next step and the DNN combined with game theory is applied for finalizing the model.

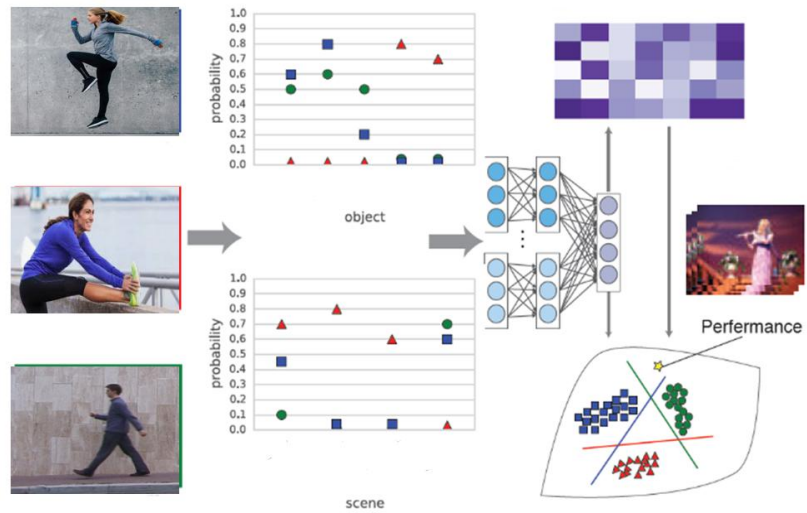

**Fig.5.** The Flowchart of the Proposed Model

### 2.3   The Scene Understanding for the Robot

Traditional robots generally only perform simple repetitive tasks. During the work process, they do not need to know working environment and environmental changes.

The operations they make are not directly related to the process and environment. Today, with more intelligent requirements for robots, robots need to obtain position data in the working environment and dynamically identify the working environment and its changes in real time. At the present, constructing the surrounding environment through vision-based scene recognition and understanding is the key technology to improve its intelligence level, and it is also the first step in the intelligentization of the service robots. Based on the DNN assisted scene recognition, it may carry on the characteristic analysis and the training to the scene, the characteristic leaks labelling to the earlier period recognition process in, labels by mistake carries on labels, and unceasingly carries on enhancement processing to the sample which forms the new dominant character collection. After the robotic control system based on deep reinforcement learning is captured by the robotic sensor camera, the pixel-level image is formed to complete the acquisition of the original image.

The multi-step deep neural network system processes the image and outputs the intermediate result, and the intermediate result is passed through the ordinary neural network. The output of robot's current decision network outputs planned actions. The robot's mechanical part drives the motor to complete the corresponding action. After the environment receives the current action, it will give a corresponding reward value by comparing the action effect and status, so that the decision network is updated and the changed state is fed back to the system, thus creating a decision cycle.

In the process of the human-robot interaction, participants can give robots encouragement and comfort through the expression of language, behavior or facial expressions, so as to achieve the purpose of guided cognitive re-evaluation. Through the proposed scene understanding model, the robot can extract the signal and finalize the orders.

## 3  Experiment and Verification

The good performance of the method in this paper is rooted in game theory. Since the images are considered as a set of people in the game, each person has its own set of the strategies. The degree of freedom of the model can approach infinity, which is helpful to improve registration ability and accuracy. Therefore, to test the proposed model, we conduct the experiment.

Because it uses the two modules alternately in the matching process the middle matching results. The gambling method assigns two kind of the characteristic similar measures to in the state player's price function carries on the multi-person decision-making to optimize, does the realization good decoupling, how can it be that compare in the simple target function optimizes a generality. The Figure 6 gives the testing result on the image pre-processing, it can be reflected that the proposed model can improve the image quality. Then, we test the scene understanding performance. In the experiment, the indoor scene was divided into 5 categories: living room, bathroom, bedroom, kitchen, and action. The figure 7 gives the scenarios sample. During model training, robot is controlled manually, and video images are collected at a rate of 20 frames per second under different light conditions. After the model training is completed, during the autonomous operation of the robot, the current real-time collected image is read every 3s as the key frame for the scene discrimination. The table 4 gives the simulation result.

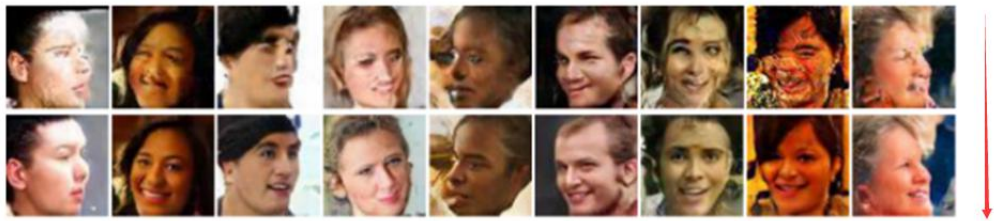

**Fig.6.** The Image Pre-processing Simulation Result

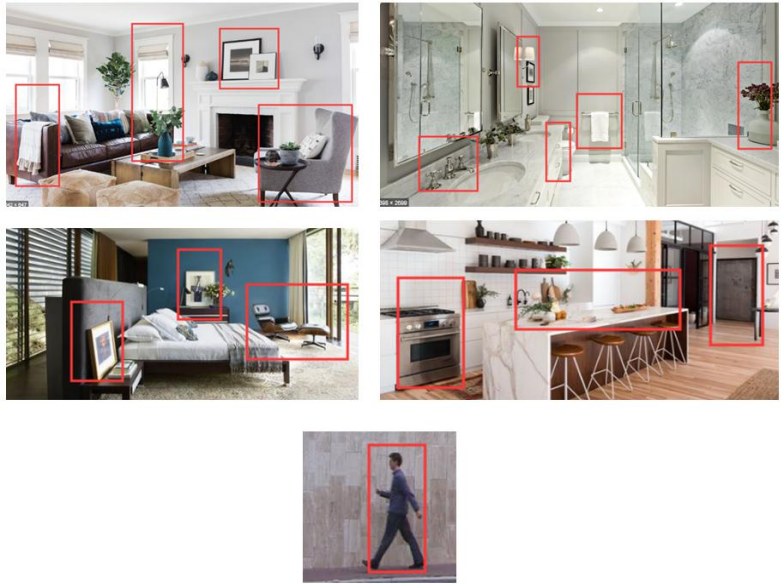

**Fig.7.** The Scenario Samples

**Table 4.** The Accuracy Testing Result for the Proposed Model

|  | Parameters | | | Final Results | |
| --- | --- | --- | --- | --- | --- |
|  | **Input Size** | **Feature Complexity Level** | **Noise Level** | **Accuracy** | **Error for Robustness** |
| **Game Level One** | 20*20 | 1 | 1 | 0.84 | **0.12±0.01** |
|  | 20*20 | 1 | 1 | 0.85 |  |
|  | 20*20 | 1 | 2 | 0.87 |  |
|  | 30*30 | 1 | 3 | 0.84 |  |
|  | 30*30 | 1 | 1 | 0.88 |  |
|  | 30*30 | 1 | 3 | 0.88 |  |
| **Game Level Two** | 40*40 | 1 | 1 | 0.90 | **0.10±0.02** |
|  | 40*40 | 1 | 1 | 0.95 |  |
|  | 40*40 | 1 | 2 | 0.90 |  |
|  | 50*50 | 1 | 3 | 0.92 |  |
|  | 50*50 | 1 | 1 | 0.93 |  |
|  | 50*50 | 1 | 3 | 0.93 |  |
| **Game Level Three** | 60*60 | 1 | 1 | 0.89 |  |
|  | 60*60 | 1 | 1 | 0.91 |  |
|  | 60*60 | 1 | 2 | 0.92 |  |
|  | 70*70 | 1 | 3 | 0.89 | **0.14±0.01** |

|  | 70*70 | 1 | 1 | 0.90 | |
|  | 70*70 | 1 | 3 | 0.91 | |
| **Game Level Four** | 80*80 | 1 | 1 | 0.87 | |
|  | 80*80 | 1 | 1 | 0.86 | |
|  | 80*80 | 1 | 2 | 0.89 | |
|  | 90*90 | 1 | 3 | 0.90 | **0.13±0.01** |
|  | 90*90 | 1 | 1 | 0.87 | |
|  | 90*90 | 1 | 3 | 0.87 | |
| **Game Level Five** | 100*100 | 1 | 1 | 0.91 | |
|  | 100*100 | 1 | 1 | 0.93 | |
|  | 100*100 | 1 | 2 | 0.91 | |
|  | 110*110 | 1 | 3 | 0.91 | **0.11±0.02** |
|  | 110*110 | 1 | 1 | 0.89 | |
|  | 110*110 | 1 | 3 | 0.90 | |

# 4  Conclusion

In this paper, we propose the scene recognition model based on the DNN and game theory with its applications in human-robot interaction. Aiming at the problem that most DNN feature extraction operators perform well in outdoor environments and degrade indoor performance, a scene recognition framework method that combines global and the salient regional features is proposed. For the better processing of the images, we use the game theory to pre-process the captured images, the quality is enhanced and the image features are extracted. The DNN is used to understand the scenarios for the vision of robot. The action of the robot can be finalized through the DNN control system. The proposed model is simulated to prove the effectiveness.